%% file: main.tex
\documentclass{article}

     \usepackage[preprint,nonatbib]{neurips_2020}
\usepackage{enumitem}
\usepackage[utf8]{inputenc} 
\usepackage[T1]{fontenc}    
\usepackage[bookmarks=false]{hyperref}
\usepackage{url}            
\usepackage{booktabs}       
\usepackage{amsfonts}       
\usepackage{nicefrac}       
\usepackage{microtype}      
\usepackage{tikz}
\usepackage{pgfplots}
\usepackage{xcolor}
\usepackage[acronym]{glossaries}

\usepackage{cleveref}
\usepackage{multirow}
\usepackage{color}
 
\newcommand{\repeatthanks}{\textsuperscript{\thefootnote}}
\usepackage{wrapfig,lipsum}

\usepackage[square,sort,comma,numbers]{natbib}

\usepackage{amsthm}

\usepackage{cleveref}
\usepackage{multirow}

\usepackage[ruled,vlined]{algorithm2e}

\usepackage{appendix}
\crefname{appsec}{Appendix}{Appendices}

\theoremstyle{definition}
\newtheorem{definition}{Definition}[section]

\newacronym{SSDL}{SSDL}{Semi-supervised deep learning}

\title{MixMOOD: A systematic approach to class distribution mismatch in semi-supervised learning using deep dataset dissimilarity measures}

\author{
 Saul Calderon-Ramirez\thanks{SCR and LO  contributed equally} \\
  Centre for Computational Intelligence\\
   De Montfort University\\
  \texttt{sacalderon@itcr.ac.cr} \\
  \And
  Luis Oala\repeatthanks\\
  Machine Learning Group\\
  Fraunhofer Heinrich Hertz Institute\\
  \texttt{luis.oala@hhi.fraunhofer.de} \\
  \AND
  Jordina Torrents-Barrena\\
  HP Inc.\\
  \texttt{jordina.torrents.barrena@hp.com} \\
  \And
  Shengxiang Yang\\
  Centre for Computational Intelligence\\
  De Montfort University\\
  \texttt{syang@dmu.ac.uk} \\
  \And
  Armaghan Moemeni\\
  School of Computer Science\\
  University of Nottingham\\
  \texttt{armaghan.moemeni@nottingham.ac.uk} \\
  \And 
  Wojciech Samek\\
  Machine Learning Group\\
  Fraunhofer Heinrich Hertz Institute\\
  \texttt{wojciech.samek@hhi.fraunhofer.de} \\
  \And 
  Miguel A. Molina-Cabello\\
  Biomedic Research Institute of M\'alaga (IBIMA)\\
  University of M\'alaga\\
  \texttt{miguelangel@lcc.uma.es} \\
}

\usepackage{comment}

\newacronym{CNN}{CNN}{Convolutional Neural Network}
\newacronym{OOD}{OOD}{out of distribution}
\newacronym{IOD}{IOD}{inside of distribution}
\newacronym{DeDiM}{DeDiM}{Deep Dataset Dissimilarity Measure}
\newacronym{MixMOOD}{MixMOOD}{MixMatch approach to out of distribution data}

\begin{document}

\maketitle              
%
\begin{abstract}
In this work, we propose MixMOOD - a systematic approach to mitigate effect of class distribution mismatch in semi-supervised deep learning (SSDL) with MixMatch. This work is divided into two components: (i) an extensive out of distribution (OOD) ablation test bed for SSDL and (ii) a quantitative unlabelled dataset selection heuristic referred to as MixMOOD. In the first part, we analyze the sensitivity of MixMatch accuracy under 90 different distribution mismatch scenarios across three multi-class classification tasks. These are designed to systematically understand how OOD unlabelled data affects MixMatch performance. In the second part, we propose an efficient and effective method, called deep dataset dissimilarity measures (DeDiMs), to compare labelled and unlabelled datasets. The proposed DeDiMs are quick to evaluate and model agnostic. They use the feature space of a generic Wide-ResNet and can be applied prior to learning. Our test results reveal that supposed semantic similarity between labelled and unlabelled data is not a good heuristic for unlabelled data selection. In contrast, strong correlation between MixMatch accuracy and the proposed DeDiMs allow us to quantitatively rank different unlabelled datasets \textit{ante hoc} according to expected MixMatch accuracy. This is what we call MixMOOD. Furthermore, we argue that the MixMOOD approach can aid to standardize the evaluation of different semi-supervised learning techniques under real world scenarios involving out of distribution data.

\end{abstract}
\section{Introduction}
\label{sec:intro}
Training an effective deep learning solution typically requires a considerable amount of labelled data. In specific application domains such as medicine, high quality labelled data can be expensive to obtain leading to scarcely labelled data settings \cite{balki2019sample,cheplygina2019not}. Several approaches have been developed to address this data constraint including data augmentation, transfer, weakly and semi-supervised learning, among others \cite{wang2017effectiveness,zhou2018brief,weiss2016survey}. Semi-supervised learning is an attractive approach for learning problems where little labelled data is available. It leverages the use of unlabelled data which is often cheap to obtain \cite{van2020survey}. Formally, in a semi-supervised setting both labelled and unlabelled datasets are used. Labelled observations $X_{l}=\left\{ \boldsymbol{x}_{1},\ldots,\boldsymbol{x}_{n_{l}}\right\}$ and their corresponding labels $Y_{l}=\left\{ y_{1},\ldots,y_{n_{l}}\right\} $ make up the labelled dataset $S_l$. The set of unlabelled observations is represented as $X_{u}=\left\{ \boldsymbol{x}_{1},\ldots,\boldsymbol{x}_{n_{u}}\right\}$, therefore $S_u = X_u$.  \gls{SSDL} approaches can be grouped into pre-training \cite{doersch2015unsupervised}, self-training or pseudo-labelled \cite{dong2018tri} and regularization based. Regularization techniques include generative based approaches, along consistency loss term and graph based regularization \cite{cheplygina2019not}. A detailed survey on semi-supervised learning can be found in \cite{van2020survey}.
The practical implementation of \gls{SSDL} techniques in different application domains has been however limited, given its  moderate success in real-world settings \cite{oliver2018realistic}. In \cite{oliver2018realistic}, authors call for more systematic and realistic evaluation of \gls{SSDL} approaches. A class distribution mismatch between labelled and unlabelled data is among the most important aspects highlighted by the authors. The following example illustrates this problem. We can train a \gls{CNN} to classify between pneumonia ill and healthy patients using chest X-ray images, as for example seen in \cite{zech2018variable}. The labelled dataset $S_l$ can include a limited number of observations for each class. However, the unlabelled dataset $S_u$ can include observations of patients with other lung pathologies. These observations correspond to what is known as \gls{OOD} data \cite{liang2018enhancing} which can potentially harm the performance of a \gls{SSDL} solution \cite{oliver2018realistic}.

\subsection{Problem statement}\label{sec:problem_statement}
The central premise of this work is the quantitative impact assessment of class distribution mismatch between labelled and unlabelled data on \gls{SSDL}. This notion stipulates that a mismatch negatively affects the accuracy of models trained with \gls{SSDL} algorithms \cite{oliver2018realistic}. A mismatch occurs when the unlabelled data contains observations that do not correspond to any of the classes present in the labelled data. It is not clear though what exactly the effect is when this mismatch occurs: does it always harm the model accuracy? Does it help to use unlabelled data that is, supposedly, semantically more similar to the labelled data? And if certain unlabelled datasets indeed harm accuracy of \gls{SSDL} trained models, is there a reliable way to select the unlabelled data in an informed way \textit{ante hoc}? In order to establish points of reference for what follows we adopt the definitions below. Given a dataset $S_1$ emanating from the data generating process $y=f(\boldsymbol{x})$, with $y\in\mathcal{Y}:= \{1,...,K \}$ being a set of labels, and a second dataset $S_2$ emanating from the data generating process $y'=g(\boldsymbol{x})$, with $y'\in\mathcal{Y'}:= \{1,...,K' \}$, we say that

\begin{definition}{\textit{Inside of distribution (\gls{IOD}) data:}} Dataset $S_2$ is \gls{IOD} relative to dataset $S_1$ if $f(\boldsymbol{x}) = g(\boldsymbol{x})$. In particular, we have that $\mathcal{Y} = \mathcal{Y'}$.
\end{definition}

\begin{definition}{\textit{Outside of distribution (\gls{OOD}) data:}} Dataset $S_2$ is \gls{OOD} relative to dataset $S_1$ if $f(\boldsymbol{x}) \neq g(\boldsymbol{x})$. In particular, we have that $\mathcal{Y} \neq \mathcal{Y'}$.
\end{definition}
\begin{definition}{\textit{Class distribution mismatch in \gls{SSDL}:}} A class distribution mismatch occurs if the unlabelled data $S_u$ used for \gls{SSDL} is \gls{OOD} relative to $S_l$.
\end{definition}
In practice, $f(\boldsymbol{x})$ and $g(\boldsymbol{x})$ are typically not known explicitly. Thus, given two datasets $S_1$ and $S_2$ a definite formal verification of the class distribution mismatch property is not possible. Instead, it is usually \textit{assumed} that two \textit{different} datasets, e.g. CIFAR-10 and MNIST, emanate from different data generating processes. This working definition of \gls{OOD} data follows the existing literature on class distribution mismatch in \gls{SSDL} \cite{oliver2018realistic} as well as \gls{OOD} detection in deep learning \cite{ren2019likelihood}. We adopt this working definition for the \gls{OOD} scenarios of our test bed. Note that different degrees of \gls{OOD} contamination for $S_u$ are possible as we describe in \Cref{sec:exp-ablation}.
\subsection{Contribution}
In order to address the questions outlined in Section \ref{sec:problem_statement} we present the \gls{MixMOOD}\footnote{All code and experimental scripts, with automatic download of test bed data for ease of reproduction, can be found at \url{https://github.com/peglegpete/mixmood}}. It entails the following contributions:

 \begin{itemize}[leftmargin=*]
    \item A systematic \gls{OOD} ablation test bed. We demonstrate that including \gls{OOD} data in the unlabelled training dataset for the MixMatch algorithm can yield different degrees of accuracy degradation compared to the exclusive use of \gls{IOD} data. However, in most cases, using unlabelled data with \gls{OOD} contamination still improves the results when compared to the default fully supervised configuration.
    \item Markedly, \gls{OOD} data that is supposedly semantically similar to the \gls{IOD} labelled data does not always lead to the highest accuracy gain.
    \item We propose and evaluate four \gls{DeDiM}s that can be used to rank unlabelled data according to the expected accuracy gain \textit{prior} to \gls{SSDL} training. They are cheap to compute and model-agnostic which make them amenable for practical application.
    \item Our test results reveal a strong correlation between the tested \gls{DeDiM}s and MixMatch accuracy, making them informative for unlabelled dataset selection. This leads to \gls{MixMOOD} which proposes the usage of tested \gls{DeDiM}s to select the unlabelled dataset for improved MixMatch accuracy.
\end{itemize}
\section{Related work}
\label{sec:soa}
In this work we address a combination of three overlapping problems that are often dealt with in parallel in the literature: class distribution mismatch in \gls{SSDL}, \gls{OOD} detection, and dataset dissimilarity measures.
\subsection{Class distribution mismatch in \gls{SSDL}}
As previously highlighted, in \cite{oliver2018realistic} authors call for the need of a more extensive testing of \gls{SSDL} techniques under real-world usage settings. One of them is the possible data distribution mismatch between the labelled and unlabelled training data. Real Mix was proposed \cite{nair2019realmix} as a reply to this remark, implementing a masking coefficient to \gls{OOD} data for the unlabelled dataset. The masking coefficient is used as a threshold of the softmax output of the model, discarding unlabelled data used only in the unsupervised term. The authors performed limited testing of the consequences of using \gls{OOD} unlabelled data. The tested \gls{OOD} dataset consists in the split CIFAR-10 dataset in two halves with different semantics. \cite{chensemi} explores a similar scenario. Four levels of \gls{OOD} contamination were tested. We argue for the need of testing different \gls{OOD} datasets to more extensively understand their impact on \gls{SSDL}. 

\subsection{\gls{OOD} data detection}
In the context of machine learning, \gls{OOD} data detection refers to the general problem of detecting observations that belong to a data distribution different from the distribution of the training data \cite{hendrycks2016baseline}. \gls{OOD} detection can  be considered as a generalization of outlier detection, since it considers individual and collective outliers \cite{singh2012outlier}. Further variations of the \gls{OOD} data detection problem are novel and anomaly data detection \cite{perera2019deep}, with different applications such as rare event detection and artificial intelligence safety \cite{hamaguchi2019rare,amodei2016concrete}. 
Classical  \gls{OOD} and anomaly detection methods rely on density estimation, e.g. Gaussian Mixture Models \cite{markou2003novelty}, robust moment estimation, like the Minimum Covariance Determinant method \cite{doi:10.1080/01621459.1984.10477105}, prototyping, e.g.  k-nearest neighbor algorithm \cite{markou2003novelty}, as well as kernel based variants such as Support Vector Data Description \cite{10.1023/B:MACH.0000008084.60811.49}. Also, a variety of neural network based approaches for novelty detection can be found \cite{markou2003novelty}, with a data oriented philosophy. 
With the success of deep learning, recent works have addressed the generic problem of discriminative detection of \gls{OOD} data for deep learning architectures. Discriminative \gls{OOD} detectors can be categorized in output- and feature-based. A simple \gls{OOD} detection approach is proposed in \cite{DBLP:journals/corr/HendrycksG16c}. The authors frame \gls{OOD} detection a prediction confidence. The proposed method relies on the softmax output, sampling the maximum value.
\cite{liang2018enhancing} introduced \gls{OOD} data detection in neural networks using input perturbations. A \emph{temperature} coefficient $T$ in the calculation of the softmax output and calibrated decision threshold $\delta$ for \gls{OOD} data detection. More recently, \cite{lee2018simple} argue that deep networks with softmax output layers are over-confident for inputs very different from the training data and hence propose the usage of the Mahalanobis distance in latent space. Similarly \cite{tagasovska2018frequentist} also exploit latent representations, defining what they refer to as learning certificates; neural networks that map feature vectors to zero for \gls{IOD} data. A more challenging \gls{OOD} detection setting was tested, where half of each tested dataset is used as
\gls{IOD} data, and the other half is used as \gls{OOD} data, making \gls{OOD} detection harder.
Another approach to \gls{OOD} detection is the use of generative adversarial learning. The generative model aims to approximate an implicit model of the \gls{IOD} data, as seen in \cite{schlegl2017unsupervised,pidhorskyi2018generative}. The standard datasets used in these test beds include MNIST, SVHN, LSUN, CIFAR-10, CIFAR-100 and Tiny ImageNet. We provide a tabular overview of these works and the explored pairing of \gls{IOD} and \gls{OOD} datasets in the appendix for the interested reader. In the reviewed literature, we can see how dataset selection for benchmarking \gls{OOD} detection commonly is not quantitatively assessed, making the comparison of the algorithms harder.
\subsection{Dataset dissimilarity measures}
Computing a notion of dissimilarity between two sets of points (known as shape matching \cite{memoli2011gromov}) is typically more computationally expensive than calculating the dissimilarity between a set of points and another single point. Strategies to reduce this burden are primarily centered around enriching the object space with a probability measure which helps guide attention to important areas of comparison \cite{memoli2011gromov}. When starting with raw datasets, as is typically the case when trying to decide which data to use for \gls{SSDL}, additional pre-processing or modelling steps would be necessary to employ these object matching strategies. Methods to compute dissimilarities between raw datasets are, to the best of our knowledge, rare. \cite{tatti2007distances} defines a dissimilarity measure based on the Euclidean distance between the frequency of a given feature function on two datasets, referred as the constrained measure distance. The calculation of the proposed measure can be efficiently performed through the usage of the covariance matrix of the feature function in the dataset. An optimized dataset measure is implemented for binary datasets. More recently, authors in \cite{cabitza2019wants} proposed a distance dissimilarity index based on the statistical significance difference of the distance distributions between the two datasets.  To calculate it,  each data point in the test set is matched with the  training data. After exchanging the associated observations,  changes in the topology are assessed, using the distance distribution. The confidence p-value of the difference between the two distributions is calculated and used as a dissimilarity measure. 

Note that our requirements differ from the above \gls{OOD} detection and dissimilarity measure methods: we are interested in computationally cheap, ante hoc and model agnostic quantification of the \gls{OOD} degree between two \textit{datasets}. Approaches that are computationally expensive or post hoc, that is being applied after the model has been trained, are not feasible to address class distribution mismatch before \gls{SSDL} training. Closest to our work are the \gls{OOD} detection ideas developed by \cite{ren2019likelihood}. The authors present introductory experiments on the correlation between \gls{OOD} detection and the dataset dissimilarity using a genome distance \cite{ren2018alignment}. We explore a similar comparison: the relationship between \gls{SSDL} accuracy and \gls{OOD}-\gls{IOD} dissimilarity.
\section{Proposed method and experiments}
\label{sec:experiments}

\subsection{Systematic \gls{OOD} ablation study}
\label{sec:exp-ablation}
\subsubsection{\gls{SSDL} setup}
The basis for all \gls{SSDL} experiments in this paper is the MixMatch algorithm, a state of the art \gls{SSDL} method \cite{berthelot2019mixmatch}. MixMatch estimates  pseudo-labels for unlabelled data $X_u$, and also implements an unsupervised regularization term. Pseudo-label $\widehat{\boldsymbol{y}}{}_{j}$ estimation is performed with the average model output of a transformed input $x_{j}$, with $K$ number of different transformations.  The pseudo-labels $\widehat{\boldsymbol{y}}$ are further sharpened with a temperature parameter $\rho$. To further augment the data using both labelled and unlabelled samples, MixMatch makes use of the MixUp algorithm by \cite{zhang2017mixup} which builds linear interpolations between labelled and unlabelled observations.
For  supervised and semi-supervised loss functions,  the cross-entropy  and the Euclidean distance, are used, respectively. The regularization coefficient  $\gamma$ controls the direct influence on unlabelled data. Unlabelled data also influences the labelled data term since unlabelled data is used also to artificially augment the dataset with the Mix Up algorithm. This loss term is used at training time, for testing, a regular cross entropy loss is implemented. We documented a detailed description of the MixMatch algorithm in the appendix along with all hyperparameters used throughout the experiments. Both follow the reference and values recommended in \cite{berthelot2019mixmatch}. 

\subsubsection{\gls{SSDL} with \gls{OOD} data ablation test bed}
\label{sec:testbed}
\input{tables/results-combined}
To assess the effect of \gls{OOD}  unlabelled data $S_u$ on the accuracy of \gls{SSDL} models trained with MixMatch, we construct an ablation test bed with four variables: base data $S_{\textrm{IOD}}$ which constitutes the original task to be learned, the type of \gls{OOD} data $T_{\textrm{OOD}}$, the \gls{OOD} data source $S_{\textrm{u,OOD}}$, the relative amount of \gls{OOD} data among the unlabelled data $\%_{\textrm{u,OOD}}$, and the amount $n_l$ of labelled observations. 
Each of the four axes is explored by varying only one of the variables at a time while keeping the others constant. This allows us to isolate the effect of the individual variables. We consider three configurations for $S_{\textrm{IOD}}$ comprising MNIST, CIFAR-10 and FashionMNIST. A total of three configurations for $T_{\textrm{OOD}}$ (\textit{half-half (HH), similar (Sim)} and \textit{different (Dif)}) are tested. We derived the possible types of \gls{OOD} data from the existing literature cited in \Cref{sec:soa}. In the \textit{half-half} setting half of the classes and associated inputs are taken to be the $S_{\textrm{IOD}}$ data whereas the other half of classes are taken to be the $S_{\textrm{u,OOD}}$ data. \textit{Similar} is a $S_{\textrm{u,OOD}}$ dataset that is supposedly semantically related to $S_{\textrm{IOD}}$, e.g. MNIST and SVHN. \textit{Different} is  a $S_{\textrm{u,OOD}}$ dataset that is supposedly semantically unrelated to $S_{\textrm{IOD}}$, e.g. MNIST and Tiny ImageNet.
There are five configurations for $S_{\textrm{u,OOD}}$ as explained above: the other half \textit{(OH)}, a similar dataset, and three different datasets including two noise baselines. They include Street View House Numbers \textit{(SVHN)}, Tiny ImageNet \textit{(TI)}, Gaussian noise \textit{(G)}, salt and pepper noise \textit{(SAP)} and Fashion Product \textit{(FP)}. Please see \Cref{tab:half-half-results} for the per task pairings. Each configuration represents a multi-class classification task with $|\mathcal{Y}|=5$, that is a random subset of half of the classes of base data $S_{\textrm{IOD}}$.  An overview for all the datasets used in our experiments can be found in the appendix. Finally, we vary the relative amount of \gls{OOD} data $\%_{\textrm{u,OOD}}$ between 0, 50 and 100 as well as the amount of labelled datapoints $n_l$ between 60, 100 and 150. Note that for each result entry you can see in \Cref{tab:half-half-results} we performed ten experimental runs and report the accuracy mean and variance of the models performing best on the test data from each run, as overfitting is very likely to happen with a low $n_l$. For each run we sampled a disjunct subset of data from $S_{\textrm{IOD}}$ and $S_{\textrm{u,OOD}}$ to obtain the required number of labelled $n_l$ and unlabelled $n_u$ samples for the run. Descriptive statistics (mean and standard deviation) for standardization of the neural networks inputs were only computed from these subsets to keep the simulation realistic and not use any information from the global training data. All other parameters (number of unlabelled observations $n_u=3000$, neural network architecture , the set of optimization hyperparameters, number of training epochs) are kept constant across all experiments to enable direct comparison with respect to the variable parameters of the system and not to  achieve state of the art performance with MixMatch on the given data. Note that it is possible to extend the test bed to other effects of interest. Some of these ideas we address at the end in \Cref{sec:discussion}. A description of all hyperparameters and the computing infrastructure used for all experiments as well as their approximate runtimes are documented in the appendix.
\subsection{Proposed method: \gls{MixMOOD} ante hoc ranking of $S_{\textrm{u,OOD}}$ benefit}
\label{sec:exp-distances}
\input{tables/distances}

In this experiment we compute the proposed \gls{DeDiM}s between the inputs of the \gls{IOD} labeled data and the inputs of the \gls{OOD} unlabelled data. In addition, we compute the correlations between the distance measures and \gls{SSDL} performance under the different \gls{OOD} configurations from the ablation experiments before. The motivation behind these distance experiments is to validate whether the measures can be used to rank different $S_{\textrm{u,OOD}}$ prior to \gls{SSDL} learning according to their expected benefit for the resulting model accuracy. We refer to this as \gls{MixMOOD}.
\subsubsection{Deep dataset dissimilarity measures}
\label{sec:distances}
In this work we implement a set of \gls{DeDiM}s. They make use of dataset subsampling, as image datasets are usually large, following a sampling approach for comparing two populations, as seen in \cite{krzanowski2003non}. We compute the dissimilarity measures in the feature space of a generic Wide-ResNet pre-trained on ImageNet, making our proposed approach agnostic to the \gls{SSDL} model to be trained. This enables an evaluation of the unlabelled data before training the \gls{SSDL} model. The proposed measures in this work are meant to be simple and quick to evaluate with practical use in mind. We propose and test the implementation of two Minkowski based dissimilarity measures, $d_{\ell_{2}}\left(S_{a},S_{b},\tau, \mathcal{C}\right)$ and $d_{\ell_{1}}\left(S_{a},S_{b},\tau, \mathcal{C}\right)$, corresponding to the Euclidean and Manhattan distances, respectively, between two datasets $S_a$ and $S_b$. Additionally, we implement and test two non-parametric density based distances; Jensen-Shannon ($d_{\textrm{JS}}$) and cosine distance ($d_{C}$). 
For all the proposed dissimilarity measures, the parameter $\tau_c$ defines the sub-sample size used to compute the dissimilarity between the two datasets $S_a$ and $S_b$ and $\mathcal{C}$ the total number of samples to compute the mean sampled dissimilarity measure. The general procedure for all the implemented distances is as follows. 

\begin{itemize}[leftmargin=*]
\item We randomly sub-sample each one of the  datasets $S_a$ and $S_b$, with a sample size of $\tau$, creating the sampled datasets $S_{a,\tau}$ and $S_{b,\tau}$.

\item We transform an input observation $\boldsymbol{x}_j\in S_i$, with $\boldsymbol{x}_j \in \mathbb{R}^{n}$, with $n$  the dimensionality of the input space, using the feature extractor $f$, yielding $ \boldsymbol{h}_{j}=f\left(\boldsymbol{x}_{j}\right)$.

Where $\boldsymbol{h}_{i}  \in \mathbb{R}^{n'}$ is the feature vector of $n'$  dimensions, with $n' < n$. For instance, the implemented feature extractor uses the Wide-ResNet architecture, extracting $n'=512$ features. This yields the two feature sets $H_{a,\tau}$ and $H_{b,\tau}$

\end{itemize}

For the Minkowski based distances $d_{\ell_{2}}\left(S_{a},S_{b},\tau, \mathcal{C}\right)$, $d_{\ell_{1}}\left(S_{a},S_{b},\tau, \mathcal{C}\right)$, we perform the following steps for the sets of features obtained in the previous description $H_{a,\tau}$ and $H_{b,\tau}$:

\begin{itemize}[leftmargin=*]
    \item For each element in $h_j \in H_{a,\tau}$, find the closest element $h_k \in H_{b,\tau}$, using the $\ell_{p}$  distance, for $d_{\ell_{p}}\left(S_{a},S_{b},\tau, \mathcal{C}\right)$, with $p=1$ or $p=2$ for the Manhattan and Euclidean distances, respectively:  
    $\widehat{d}_{j}=\underset{k}{\textrm{argmin}}\left\Vert \boldsymbol{h}_{j}-\boldsymbol{h}_{k}\right\Vert _{p}.$ We do this for a number of $\mathcal{C}$ samples, yielding a list of distance calculations $\widehat{d}_{1}, \widehat{d}_{2},..., \widehat{d}_\mathcal{C}$. 
    
    \item We compute a reference list of distances for the same list of samples of the dataset $S_a$ to itself (intra-dataset distance), computing $d_{\ell_{p}}\left(S_{a},S_{a},\tau, \mathcal{C}\right)$. This yields a list of reference distances $\check{d}_{1}, \check{d}_{2}, ..., \check{d}_{\mathcal{C}}$. In our case $S_a$ corresponds to the labelled dataset $S_l$, as the distance to different unlabelled datasets $S_u$ is to be computed.
    
    \item To ensure that the absolute differences between the reference and inter-dataset distances $d_{c}=\left|\widehat{d}_{c}-\check{d}_{c}\right|$ are statistically significant, we compute the $p$ significance value with a Wilcoxon test. 
    
    \item Computing the distance between two datasets $d_{\ell_{p}}\left(S_{a},S_{b},\tau, \mathcal{C}\right)$ results in the average reference substracted distance $\overline{d}$ and its corresponding confidence $p$ value. 
\end{itemize}
As for the density based distances implemented we follow a similar sub-sampling approach, with these steps:
\begin{itemize}[leftmargin=*]
    \item For each dimension $r=1,...,n'$ in the feature space, we compute the normalized histogram $p_{r,a}$, in the sample $H_{a,\tau}$. Similarly, we compute the set of  density functions $p_{r,b}$ for $r=1,...,n'$, using the observations in the sample $H_{b,\tau}$. 
    \item We compute the sum of the distances between the density functions $p_{r,a}$ and $p_{r,b}$, to yield the distance approximation for the sample $j$: 
    $\widehat{d}_{j}=\ensuremath{\sum_{r=1}^{n'}\delta_{g}\left(p_{r,a},p_{r,b}\right)}$. 
    We do this for all the $\mathcal{C}$ samples, yielding the list of inter-dataset distances: $\widehat{d}_{1}, \widehat{d}_{2},..., \widehat{d}_\mathcal{C}$. To lower the computational burden,  we assume that the dimensions are statistically independent. 
    
    \item Similar to the Minkowski distances, we compute the intra-dataset distances for the dataset $S_a$, in this context the labelled dataset $S_l$, to obtain the list of reference distances $\check{d}_{1}, \check{d}_{2}, ..., \check{d}_{\mathcal{C}}$.
    
    \item Similarly, to verify that the inter- and intra-dataset distance difference  $d_{c}=\left|\widehat{d}_{c}-\check{d}_{c}\right|$ are statistically significant, we compute the $p$ significance value with a Wilcoxon test. The distance computation yields the sample mean distance   $\overline{d}$ and its confidence value $p$.
\end{itemize}

The proposed dissimilarity measures do not hold for a mathematical metric or pseudo-metric, as the distance of an object to itself is not strictly zero (but tends to be close) and symmetry properties are not fulfilled for the sake of evaluation speed \cite{cremers2003pseudo}. Despite these relaxations we will see that these dissimilarity measures, especially the two that are density based, are an effective proxy for $S_{\textrm{u,OOD}}$ benefit.
To quantitatively measure how related the distances between $S_l$ and $S_u$ and the yielded \gls{SSDL} accuracy are, we calculate the Pearson coefficient between the distance measures and the \gls{SSDL} accuracy. This verifies the linear correlation between them. Table \ref{tab:correlations} describes the Pearson coefficient for each implemented dissimilarity measure and each \gls{SSDL} configuration. In summary, as part of \gls{MixMOOD} we propose to quantitatively rank a set of possible unlabelled datasets $S_{u,1}, S_{u,2},...,S_{u,k}$ according to a dissimilarity measure $d(S_l, S_u)$, instead of using qualitative based heuristics. In all the tests of this work, we used $n' = 512$, $\tau = 80$ and $\mathcal{C} = 30$.
\section{Results}
\label{sec:results}
Table  \ref{tab:half-half-results} shows the results of the distribution mismatch experiment described in \Cref{sec:exp-ablation}. We make a number of observations. 
First, in the majority of cases using \gls{IOD} unlabelled data or a 50-50 mix of \gls{IOD} and \gls{OOD} unlabelled data beats the fully supervised baseline. The gains range from 15\% to 25\% for MNIST, 10\% to 15\% for CIFAR-10 and 7\% to 13\% for FashionMNIST across all $S_{\textrm{u,OOD}}$ and $n_l$. As expected, in most of the cases the accuracy is degraded when including \gls{OOD} data in $S_u$, with a more dramatic hit when noisy datasets (SAP, G) are used as  \gls{OOD} data contamination.
Second, it is not always the case that $T_{\textrm{OOD}}=\textrm{HH}$, when $S_{\textrm{u,OOD}}$ is supposedly most similar to $S_{\textrm{IOD}}$, yields the best MixMatch performance. This is observed for CIFAR-10, when $n_l=100$ and $n_l=150$, where \gls{OOD} unlabelled data from Tiny ImageNet results in more accurate models than using the other half of CIFAR-10 as $S_{\textrm{u,OOD}}$. It is interesting that an $S_{\textrm{u,OOD}}$ dataset of type \textit{different} can be more beneficial than a $S_{\textrm{u,OOD}}$ dataset of type \textit{similar} which is also the case for FashionMNIST and Tiny ImageNet 
\input{tables/correlations}
versus Fashion Product at $n_l=150$. This contradicts the common heuristic that unlabelled data that appears semantically more related to the labelled data is always the better choice for \gls{SSDL}. Rather, as we demonstrate in the second set of results below, a notion of distance between labelled and unlabelled data offers a more consistent and quantifiable proxy for the expected benefit of different unlabelled datasets. 

The second set of results demonstrate the potential of using distance measures as a systematic and quantitative ranking heuristic when selecting unlabelled datasets for the MixMatch algorithm. The exact distances, as described in \Cref{sec:exp-distances}, for all \gls{OOD} configurations from the ablation study can be found in \Cref{tab:distances}. We can observe that these distances trace the accuracy results found in \Cref{tab:half-half-results}. This correlation is quantified in \Cref{tab:correlations} with the cosine based density measure $d_c$  correlating particularly well with the accuracy results of \Cref{tab:half-half-results}. Also, the p-values  are consistently lower for the density based distances, meaning that density based distances can be enjoyed with more confidence as seen in \Cref{tab:distances}. We suspect that this is related to the quantitative approximation of the distribution mismatch implemented both in the $d_{\textrm{JS}}$ and $d_{\textrm{C}}$ distances which we plan to explore in future work.
\section{Conclusions and recommendations}
\label{sec:discussion}
In this work we extensively tested the behavior of the MixMatch algorithm under various \gls{OOD} unlabelled data settings. We introduced \gls{MixMOOD}, which uses quantitative data selection heuristics, DeDiMs, to rank unlabelled datasets \textit{ante hoc} according to their expected  benefit to \gls{SSDL}. Our results lead us to the following conclusions and recommendations:

\begin{itemize}[leftmargin=*]
\item Real-world usage scenarios of \gls{SSDL} can include different degrees of \gls{OOD} data contamination: for instance, with a deep learning model trained for medical imaging analysis, unlabelled data can include images within the same domain, but capturing different pathologies not present in the labelled data. This scenario has been simulated with the \emph{half-half} setting which resulted in a subtle accuracy degradation in most cases. However, the accuracy gain obtained vis-a-vis the fully supervised baseline is still substantial, making the application of \gls{SSDL} attractive in such a setting.
\item Another plausible real-world scenario for \gls{SSDL} is the inclusion of widely available unlabelled datasets, e.g. built with web crawlers, where the domain shift can be more substantial. This scenario has been simulated with the \gls{OOD} types similar and different. We can observe that notions of semantic similarity between labelled and unlabelled dataset pairings, e.g. (MNIST-SVHN) or (FashionMNIST-Fashion Product), do not necessarily imply an \gls{SSDL} accuracy gain. Distance measures, in particular $d_C$, are an accurate and systematic  proxy for \gls{SSDL} accuracy. This is visible when comparing the accuracy and distance results of the previous pairings to (MNIST-Tiny ImageNet) and (FashionMNIST-Tiny ImageNet) which have higher accuracies and, also, surprisingly, lower measurements. 
\item Overall, the implemented \gls{DeDiM}s correlate strongly with the yielded \gls{SSDL} accuracy, in particular density based measures, recommended for its usage in \gls{MixMOOD}. This approach can be applied in \gls{SSDL} prior to learning to aid the unlabelled data selection process and mitigate the class distribution mismatch problem. The facts that they are model agnostic, simple and fast to compute make them particularly suitable for practical application in \gls{SSDL}.
\item Finally, the proposed test bed and distance measures can be used for a more systematic quantitative evaluation of \gls{SSDL} algorithms.
\end{itemize}

In future work, this test bed can also be applied to other \gls{SSDL} variants, depth-first analyses (e.g. fewer tasks with more training epochs), additional axes of test bed variables (e.g. $n_u$) and more testing around the appropriate dissimilarity measures parameters.
\newpage
\section*{Broader Impact}

We note that this study constitutes a breadth-first ablation exploration. Our goal was to present a comprehensive test bed to better understand the class distribution mismatch problem and provide a more systematic and quantitative approach for mitigating it. So far these results are limited to the MixMatch algorithm. The extension and usage of the proposed method needs further assessment, specially for its use in particular domains, where social and human costs are considerable. 
\begin{ack}
The authors would like to thank S{\"o}ren Becker for detailed feedback on a first draft of this paper.
\end{ack}
\bibliographystyle{unsrt}
\bibliography{references/mendeley,references/related_work,references/luis_refs,references/sauls_references}

\newpage
\begin{appendices}
\section{MixMatch: Detailed description of the SSL algorithm used in this paper}
In MixMatch, the consistency loss term minimizes the distance of the pseudo-labels and the model predictions over the unlabelled dataset $X_{u}$. Pseudo-label $\widehat{\boldsymbol{y}}{}_{j}$ estimation is performed with the average model output of a transformed input $x_{j}$, with $K$ number of different transformations.  $K=2$ is advised in \cite{berthelot2019mixmatch}. The estimated pseudo-labels $\widehat{\boldsymbol{y}}$ might be too \textit{unconfident}. To tackle this, pseudo-label sharpening is performed with a temperature $\rho$.  The dataset with the estimated and sharpened pseudo-labels was defined as $\widetilde{S}_{u}=\left(X_{u},\widetilde{Y}\right)$, with $\widetilde{Y}=\left\{ \widetilde{\boldsymbol{y}}_{1},\widetilde{\boldsymbol{y}}_{2},\ldots,\widetilde{\boldsymbol{y}}_{n_{u}}\right\}$.

Data augmentation is a key aspect in semi-supervised deep learning as found in \cite{berthelot2019mixmatch}. To further augment data using both labelled and unlabelled samples, they implemented the Mix Up algorithm developed in \cite{zhang2017mixup}. Linear interpolation of a mix labelled observations and unlabelled (with its corresponding pseudo-labels) observations.
\begin{equation}
\left(S'_{l},\widetilde{S}'_{u}\right)=\Psi_{\textrm{MixUp}}\left(S_{l},\widehat{S}_{u},\alpha\right)
\end{equation}

The Mix Up algorithm creates new observations from a linear interpolation of a mix of unlabelled (with its corresponding pseudo-labels) and labelled data. More specifically, it takes two labelled (or pseudo labelled) data pairs $\left(\boldsymbol{x}_{a},y_{a}\right)$ and $\left(\boldsymbol{x}_{b},y_{b}\right)$. The Mix Up method generates a new observation and its label $\left(\boldsymbol{x}',y'\right)$ by following these steps: 

\begin{enumerate}
\item  Sample the Mix Up parameter $\lambda$ from a Beta distribution $\lambda\sim\textrm{Beta}\left(\alpha,\alpha\right)$.

\item  Ensure that $\lambda>0.5$ by making $\lambda'=\max\left(\lambda,1-\lambda\right)$.

\item  Create a new observation with a lineal interpolation of both observations: $\boldsymbol{x}'=\lambda'\boldsymbol{x}_{a}+\left(1-\lambda'\right)\boldsymbol{x}_{b}$.
\end{enumerate}

With the augmented datasets $ \left(S'_{l},\widetilde{S}'_{u}\right)$, the MixMatch training can be summarized as:
\begin{equation}
f_{\boldsymbol{w}}=T_{\textrm{MixMatch}}\left(S_{l},S_{u},\alpha,\gamma, \lambda\right)=\underset{\boldsymbol{w}}{\textrm{argmin}}\mathcal{L}\left(S,\boldsymbol{w}\right)
\end{equation}

\begin{equation}
\mathcal{L}\left(S,\boldsymbol{w}\right)=\sum_{\left(\boldsymbol{x}_{i},\boldsymbol{y}_{i}\right)\in S'_{l}}\mathcal{L}_{l}\left(\boldsymbol{w},\boldsymbol{x}_{i},\boldsymbol{y}_{i}\right)+
r(t) \gamma\sum_{\left(\boldsymbol{x}_{j},\widetilde{\boldsymbol{y}}_{j}\right)\in\widetilde{S}'_{u}}\mathcal{L}_{u}\left(\boldsymbol{w},\boldsymbol{x}_{j},\widetilde{\boldsymbol{y}}_{j}\right)
\end{equation}

For  supervised and semi-supervised loss functions,  the cross-entropy $\mathcal{L}_{l}\left(\boldsymbol{w},\boldsymbol{x}_{i},\boldsymbol{y}_{i}\right)=\delta_{\textrm{cross-entropy}}\left(\boldsymbol{y}_{i},f_{\boldsymbol{w}}\left(\boldsymbol{x}_{i}\right)\right)$ and the Euclidean distance $\mathcal{L}_{u}\left(\boldsymbol{w},\boldsymbol{x}_{j},\widetilde{\boldsymbol{y}}_{j}\right)=\left\Vert \widetilde{\boldsymbol{y}}_{j}-f_{\boldsymbol{w}}\left(\boldsymbol{x}_{j}\right)\right\Vert$, are usually implemented, respectively. The regularization  $\gamma$ controls the direct influence on unlabelled data. Since in the first epochs, unlabelled data based predictions are unreliable, the function $r(t) = t/\rho$ increases the unsupervised term contribution as the number of epochs progress.  The coefficient $\rho$ is referred as the rampup coefficient.  Unlabelled data also influences the labelled data term $\mathcal{L}_{l}$, since unlabelled data is used also to artificially augment the dataset with the Mix Up algorithm. This loss term is used at training time, for testing, a regular cross entropy loss is implemented.
\newpage
\section{MixMOOD pseudocode description}
Note that based on our empirical results we recommend the use of density based deep dissimilarity measures, in particular cosine distance, as these displayed the best correlation with MixMatch accuracy. 
\begin{algorithm}[h]
\SetAlgoLined
\KwIn{A list of unlabelled datasets $S_{u_1}, S_{u_2}, ..., S_{u_k}$}

For each unlabelled dataset $S_{u_i}$ do:
    
\begin{enumerate}
    \item Randomly sub-sample each one of the  datasets $S_l$ and $S_{u_i}$, with a sample size of $\tau$, creating the sampled datasets $S_{l,\tau}$ and $S_{u_i,\tau}$.
    \item Transform all input observations in the two samples $\boldsymbol{x}_j\in S_i$, with $\boldsymbol{x}_j \in \mathbb{R}^{n}$, with $n$  the dimensionality of the input space, using the feature extractor $f$, yielding $ \boldsymbol{h}_{j}=f\left(\boldsymbol{x}_{j}\right)$. Where $\boldsymbol{h}_{i}  \in \mathbb{R}^{n'}$ is the feature vector of $n'$  dimensions, with $n' < n$. For instance, the implemented feature extractor uses the Wide-ResNet architecture, extracting $n'=512$ features. This yields the two feature sets $H_{l,\tau}$ and $H_{u_i,\tau}$
    
    \item For each dimension $r=1,...,n'$ in the feature space, compute the normalized histogram $p_{r,l}$, in the sample $H_{l,\tau}$. Similarly, we compute the set of  density functions $p_{u_i,b}$ for $r=1,...,n'$, using the observations in the sample $H_{u_i,\tau}$. 
    \item Compute the sum of the distances between the density functions $p_{r,l}$ and $p_{r,u_i}$, to yield the distance approximation for the sample $j$: 
    $\widehat{d}_{j}=\ensuremath{\sum_{r=1}^{n'}\delta_{g}\left(p_{r,a},p_{r,b}\right)}$. 
    We do this for all the $\mathcal{C}$ samples, yielding the list of inter-dataset distances: $\widehat{d}_{1}, \widehat{d}_{2},..., \widehat{d}_\mathcal{C}$. To lower the computational burden,  we assume that the dimensions are statistically independent. 
    
    \item Compute the intra-dataset distances for the dataset $S_l$, in this context the labelled dataset $S_l$, to obtain the list of reference distances $\check{d}_{1}, \check{d}_{2}, ..., \check{d}_{\mathcal{C}}$.
    
    \item Compute the $p$ significance value with a Wilcoxon test to verify that the inter- and intra-dataset distance difference  $d_{c}=\left|\widehat{d}_{c}-\check{d}_{c}\right|$ are statistically significant. The distance computation yields the sample mean distance   $\overline{d}_{u_i}$ and its confidence value $p_{u_i}$.

\end{enumerate} 
Pick the unlabelled dataset $S_{u_{\textrm{best}}}$ with the lowest distance $\overline{d}_{u_{\textrm{lowest}}}$.

\KwResult{$S_{u_{\textrm{best}}}$ the unlabelled dataset to yield the best accuracy for MixMatch}

 \caption{MixMOOD for unlabelled dataset selection}
\end{algorithm}
\newpage
\section{Hyperparameters}
\subsection{Global}
\begin{table}[h]
\caption{Global hyperparameters which are kept constant throughout all experiments. This was done in order to isolate the effects of the changing OOD data configurations.}
\centering
\begin{tabular}{ccc}
\toprule
\textbf{Description} & \textbf{Name in code}             & \textbf{Value} \\
\midrule
Model architecture used in all tasks                & \url{MODEL}& wide\_resnet           \\
Number of training epochs            & \url{EPOCHS} & $50$          \\
Batch size & \url{BATCH_SIZE} & $16$     \\ 
Learning rate & \url{LR} & $0.0002$\\
Weight decay & \url{WEIGHT_DECAY} & $0.0001$\\
Rampup coefficient & \url{RAMPUP_COEFFICIENT} & $3000$\\
Optimizer & - & Adam with\\
& & 1-cycle policy \cite{smith2018disciplined}\\
\bottomrule
\end{tabular}
\vspace{1mm}
\label{table:globalhyper}
\end{table}
\subsection{MixMatch}
\begin{table}[h]
\caption{MixMatch hyperparameters. All parameters were chosen following the recommendations by \cite{berthelot2019mixmatch}.}
\centering
\begin{tabular}{cccc}
\toprule
\textbf{Symbol} & \textbf{Description} & \textbf{Name in code}             & \textbf{Value} \\
\midrule
$K$                & Number of augmentations                & \url{K_TRANSFORMS}& $2$            \\
$T$                & Sharpening temperature            & \url{T_SHARPENING} & $0.5$          \\
$\alpha$           & Parameter for the Beta distribution & \url{ALPHA_MIX} & $0.75$     \\ 
$\gamma$ & Gamma for the loss weight & \url{GAMMA_US} & $25$ \\
- & Whether to use balanced (5) & \url{BALANCED} & $5$\\
 &  or unbalanced (-1) loss for MixMatch & & \\
\bottomrule
\end{tabular}
\vspace{1mm}
\label{table:mixmatchhyper}
\end{table}

\section{Dataset descriptions}

If you wish to reproduce any of the experiments datasets are automatically downloaded by the experiment script \url{ood_experiment_at_scale_script.sh} for your convenience based on which experiment you choose to run.

An overview of the different datasets can be found below. Note that we used the training split of each dataset as the basis to construct our own training and test splits for each experimental run.

\begin{table}[h]
\caption{Information on the datasets used in the experiments. \textbf{Format} specifies the format the image files were provided in, $\mathbf{d}$ specifies the size of the images, $\mathbf{N}$ specifies the number of samples in the dataset, $\mathbf{|\mathcal{Y}|}$ specifies the number of classes in the dataset, \textbf{Relative class distribution} specifies the relative class distribution in the dataset.}
\centering
\begin{tabular}{cccccc}
\toprule
\textbf{Dataset} & \textbf{Format} & $\mathbf{d}$  & $\mathbf{N}$ & $\mathbf{|\mathcal{Y}|}$ & \textbf{Relative class distribution} \\
\midrule
\textbf{MNIST}\cite{lecun1998gradient} & \url{.jpg} & $28 \times 28$ & 42,000 & 10 & Uniform\\
\multirow{2}{*}{\textbf{SVHN}\cite{netzer2011reading}} &\url{.png} & $32 \times 32$ & 73,557 & 10 & 0.07/0.19/0.14/0.12/0.1/ \\
& & & & & 0.09/0.08/0.07/0.07/0.07 \\
\textbf{Tiny ImageNet}\cite{deng2009imagenet} & \url{.jpg} & $64 \times 64$ & 100,000 & 200 & Uniform\\
\textbf{CIFAR-10}\cite{krizhevsky2009learning} & \url{.jpg} & $32 \times 32$ & 50,000 & 10 & Uniform\\
\textbf{FashionMNIST}\cite{xiao2017} & \url{.png} & $28 \times 28$& 60,000 & 10 & Uniform\\
\textbf{Fashion Product}\cite{fashionproduct} & \url{.jpg} & $60 \times 80$ & 44,441 & 5 & 0.48/0.25/0.21/0.05/0.01 \\
\textbf{Gaussian} & \url{.png} & $224 \times 224$& 20,000 & NA & NA\\
\textbf{Salt and Pepper} & \url{.png} & $224 \times 224$& 20,000 & NA & NA\\
\bottomrule
\end{tabular}
\vspace{1mm}
\label{table:dataset}
\end{table}
The Gaussian and Salt and Pepper datasets were created with the following parameters: a variance of 10 and mean 0 for the Gaussian noise, and an equal Bernoulli probability for 0 and 255 pixels, in the case of the Salt and Pepper noise. 

\subsection{Preprocessing}
Each data point was preprocessed in the following way. After a subset of labelled and unlabelled data for an experimental run had been constructed the means and standard deviations (one pair for labelled data, one pair for unlabelled data) were calculated for this specific subset. Then, the labelled and unlabelled inputs were standardized by subtracting the respective mean and dividing by the respective standard deviation.

In addition, in situations when the size of the unlabelled images differed from the size of the labelled images up- or downsampling was used to align the unlabelled image size.

\section{Existing OOD detection methods and IOD-OOD data pairings}
\begin{table}[h]
\caption{OOD test benchmarks for different techniques. Datasets with * were randomly cut by half for in-distribution training labelled data and the other half was used as OOD unlabelled data. The table reveals how arbitrary different testbeds have been used for benchmarking OOD detection algorithms. IOD-OOD dataset pairs are indicated by number pairs in the table.}
\label{table:datasets_ood}
\centering
\begin{tabular}{ccc}
\toprule
\textbf{Method name}                                                                   & \textbf{IOD data} & \textbf{OOD data} \\
\midrule
\multirow{5}{*}{Max. value of Softmax layer  \cite{DBLP:journals/corr/HendrycksG16c}}  & CIFAR-10 $^{1}$                   & SUN$^{1,2}$                           \\
                                                                                       & CIFAR-100 $^{2}$                  & Gaussian $^{1,2}$                     \\
                                                                                       & MNIST $^{3}$                      & Omniglot $^{3}$                       \\
                                                                                       &                                   & notMNIST$^{3}$                        \\
                                                                                       &                                   & Uniform noise$^{3}$                   \\ \midrule
\multirow{4}{*}{Inhibited Softmax\cite{mozejko2018inhibited}}                          & CIFAR-10$^{1}$                    & SVHN$^{1}$                            \\
                                                                                       & MNIST$^{2}$                        & LFW-A$^{1}$                           \\
                                                                  &              & notMNIST$^{2}$                        \\
                                                                  &             & Omniglot$^{2}$                        \\ \midrule
\multirow{5}{*}{ODIN \cite{liang2018enhancing}}                                        & CIFAR-10$^{1}$                    & TinyImageNet$^{1,2}$                  \\
                                                                                       & CIFAR-100$^{2}$                   & LSUN$^{1,2}$                          \\
                                                                                       &                                   & iSUN$^{1,2}$                          \\
                                                                                       &                                   & Uniform$^{1,2}$                       \\
                                                                                       &                                   & Gaussian$^{1,2}$                      \\ \midrule
\multirow{4}{*}{Epistemic Uncertainty Estimation \cite{tagasovska2018frequentist}}     & CIFAR *$^{1}$                     & CIFAR*$^{1}$                          \\
                                                                                       & FashionMNIST*$^{2}$               & FashionMNIST*${^2}$  \\
                                                                                       & SVHN*$^{3}$                       & SVHN*$^{3}$                           \\
                                                                                       & MNIST*$^{4}$                      & MNIST*$^{4}$                          \\ \hline
\multirow{4}{*}{Mahalanobis latent distance \cite{lee2018simple}} & CIFAR-10$^{1}$                    & SVHN$^{1,2}$                          \\
                                                                                       & CIFAR-100$^{2}$                   & CIFAR-10$^{3}$                        \\
                                                                                       & SVHN$^{3}$                        & TinyImageNet$^{1,2,3}$                \\
                                                                                       &                                   & LSUN$^{1,2,3}$     \\                  
\bottomrule
\end{tabular}
\end{table}
\section{Code archive and computing infrastructure}
All training and evaluation code can be found at \url{https://github.com/peglegpete/mixmood}. Software dependencies are specified in the \url{requirements.txt} file in the same archive. We use the \url{mlflow} framework for experiment management and reproduction. After experiments have been completed you can extract results from all runs using the analysis scripts provided in the archive.

Experiments were run on three machines. Machine 1 has one 12GB NVIDIA TITAN X GPU, 24 Intel(R) Xeon(R) CPU E5-2687W v4 @ 3.00GHz and 32GB RAM. Machine 2 has four 16GB NVIDIA T4 GPUs, 44 CPUs from the Intel Xeon Skylake family and 150GB RAM. Machine 3 has one 12GB NVIDIA TITAN V GPU, 24 Intel(R) Xeon(R) E5-2620 0 @ 2.00GHz CPU and 32GB RAM. 

Experimental runs were parallelized using the ampersand option in \url{bash} executing 10 runs in parallel on a single GPU. With the current code base this requires up to 10 CPUs per GPU as well as approximately 25GB RAM per GPU. With this setup a single training epoch of 10 parallel experimental runs should last between 2 and 4 minutes per GPU, depending on which type of GPU is used. 
\end{appendices}

\end{document}

%% file: tables/results-combined.tex
\begin{table}[t]
\caption{Results for the class distribution mismatch experiment. Each result entry in the table represents the mean and variance of accuracy across ten random experimental runs per entry. For a detailed description of symbols and the experiment see \Cref{sec:testbed}.}
\label{tab:half-half-results}
\tiny
\centering
\begin{tabular}{ccccccc}
\toprule
\multirow{2}{*}{$\mathbf{S_{IOD}}$} & \multirow{2}{*}{$\mathbf{T_{OOD}}$} & \multirow{2}{*}{$\mathbf{S_{uOOD}}$} & \multirow{2}{*}{$\mathbf{\%_{uOOD}}$} & \multicolumn{3}{c}{$\mathbf{n_l}$}\\

&  & &  & 60 & 100 & 150\\

\midrule

\multirow{12}{*}{\rotatebox[origin=c]{90}{\textbf{MNIST}}} 
& \multicolumn{3}{c}{Fully supervised baseline} & $0.457\pm0.108$ & $0.559\pm 0.125 $ & $0.645 \pm0.101$ \\

 & \multicolumn{3}{c}{SSDL baseline (no OOD data)} & $\mathbf{0.704 \pm 0.096}$ & $\mathbf{0.781\pm 0.065}$ & $\mathbf{0.831 \pm0.0626}$\\

 \cline{2-7}
 
 & \multirow{2}{*}{HH} & \multirow{2}{*}{OH} & 50 & $0.679\pm0.108$ & $0.769\pm0.066$ & $0.802\pm0.054$\\
 & & & 100 & $0.642 \pm 0.111$ & $0.746 \pm0.094$ & $0.798 \pm0.07$\\

 \cline{2-7}
 & \multirow{2}{*}{Sim} & \multirow{2}{*}{SVHN} & 50 & $0.637 \pm0.098$ & $0.745 \pm0.081$ & $0.801 \pm0.0699$\\
 
 & & & 100 & $0.482 \pm0.113$ & $0.719 \pm0.058$ & $0.765 \pm0.072$\\
 
  \cline{2-7}
 
  & \multirow{6}{*}{Dif}& \multirow{2}{*}{TI} & 50 & $0.642 \pm0.094$ & $0.739 \pm 0.074$ & $0.809 \pm0.066$\\
  
  & & & 100 & $0.637 \pm0.097$ & $0.732\pm 0.074$ & $0.804 \pm0.071$\\

 \cline{3-7}
 
  & & \multirow{2}{*}{G} & 50 & $0.606 \pm0.0989$ & $0.713 \pm0.087$ & $0.786\pm 0.065$\\
  
  & & & 100 & $0.442 \pm0.099$ & $0.461 \pm0.073$ & $0.542\pm 0.062$\\

   \cline{3-7}
 
  & & \multirow{2}{*}{SAP} & 50 & $0.631 \pm0.102$ & $0.735 \pm0.082$ & $0.813 \pm0.057$\\
  
  & & & 100 & $0.48\pm0.0951$ & $0.524 \pm0.09$ & $0.563 \pm0.095$\\
 
\hline

\multirow{12}{*}{\rotatebox[origin=c]{90}{\textbf{CIFAR-10}}} & \multicolumn{3}{c}{Fully supervised baseline} & $0.380\pm0.024$ & $0.445\pm0.042$ & $0.449\pm0.022$\\

 &  \multicolumn{3}{c}{SSDL baseline (no OOD data)} & $\mathbf{0.453\pm0.046}$ & $0.474\pm0.019$ & $0.501\pm0.033$\\

 \cline{2-7}

& \multirow{2}{*}{HH} & \multirow{2}{*}{OH} & 50 & $0.444\pm0.040$ & $0.472\pm0.039$ & $0.525\pm0.050$\\ 
 & & & 100 & $0.443\pm0.023$ & $0.472\pm0.047$ & $0.499\pm0.054$ \\
 \cline{2-7}
 & \multirow{2}{*}{Sim} & \multirow{2}{*}{TI} & 50 & $0.435\pm0.054$ & $0.473\pm0.039$ & $\mathbf{0.543\pm0.040}$\\ 
 & & & 100 & $0.417\pm0.020$ & $\mathbf{0.480\pm0.039}$ & $0.498\pm0.042$ \\
 
 \cline{2-7}
 
  & \multirow{6}{*}{Dif} & \multirow{2}{*}{SVHN} & 50 & $0.419\pm0.027$ & $0.464\pm0.044$ & $0.469\pm0.056$\\ 
 & & & 100 & $0.385\pm0.034$ & $0.418\pm0.035$ & $0.440\pm0.046$ \\
 
 \cline{3-7}
 
  & & \multirow{2}{*}{G} & 50 & $0.409\pm0.047$ & $0.454\pm0.048$ & $0.491\pm0.032$\\ 
 & & & 100 & $0.297\pm0.029$ & $0.306\pm0.034$ & $0.302\pm0.038$ \\
 
 \cline{3-7}
 
  & & \multirow{2}{*}{SAP}& 50 & $0.438\pm0.029$ & $0.455\pm0.037$ & $0.485\pm0.034$\\ 
 & & & 100 & $0.236\pm0.031$ & $0.246\pm0.032$ & $0.232\pm0.022$ \\
  
  \hline

\multirow{12}{*}{\rotatebox[origin=c]{90}{\textbf{FashionMNIST}}} & \multicolumn{3}{c}{Fully supervised baseline} & $0.571\pm0.073$ & $0.704\pm0.066$ & $0.720\pm0.093$\\

 & \multicolumn{3}{c}{SSDL baseline (no OOD data)} & $\mathbf{0.715\pm0.049}$ & $\mathbf{0.760\pm0.044}$ & $0.756\pm0.069$\\
 \cline{2-7}

& \multirow{2}{*}{HH} & \multirow{2}{*}{OH} & 50 & $0.714\pm0.049$ & $0.721\pm0.104$ & $0.765\pm0.053$\\ 
 & & & 100 & $0.660\pm0.061$ & $0.711\pm0.090$ & $0.747\pm0.061$ \\
 \cline{2-7}
 & \multirow{2}{*}{Sim} & \multirow{2}{*}{FP} & 50 & $0.707\pm0.039$ & $0.724\pm0.030$ & $0.778\pm0.078$\\ 
 & & & 100 & $0.546\pm0.101$ & $0.542\pm0.099$ & $0.540\pm0.105$ \\
 
 \cline{2-7}
 
  & \multirow{6}{*}{Dif}& \multirow{2}{*}{TI} & 50 & $0.690\pm0.065$ & $0.745\pm0.093$ & $0.792\pm0.058$\\ 
 & & & 100 & $0.690\pm0.073$ & $0.728\pm0.066$ & $\mathbf{0.794\pm0.056}$ \\
  \cline{3-7}
  & & \multirow{2}{*}{G} & 50 & $0.644\pm0.061$ & $0.689\pm0.075$ & $0.755\pm0.055$\\ 
 & & & 100 & $0.352\pm0.025$ & $0.366\pm0.065$ & $0.361\pm0.057$ \\
  \cline{3-7}
  & & \multirow{2}{*}{SAP} & 50 & $0.671\pm0.072$ & $0.708\pm0.095$ & $0.729\pm0.088$\\ 
 & & & 100 & $0.276\pm0.069$ & $0.297\pm0.046$ & $0.283\pm0.059$\\
 \bottomrule
\end{tabular}
\end{table}

%% file: tables/distances.tex
\begin{table}[t]
\caption{Distance measures between the labelled and unlabelled datasets $S_l$ and $S_u$. Numbers in italics correspond to results with $p>0.5$ for the Wilcoxon test. For a detailed description of symbols and the experiment see \Cref{sec:distances}}
\label{tab:distances}
\begin{center}
\begin{tiny}
\begin{tabular}{ccccccc} 
\toprule
$\mathbf{S_l}$     & $\mathbf{S_u}$  & $\mathbf{\%_{uOOD}}$  & $\mathbf{d_{\ell_2}}$  & $\mathbf{d_{\ell_1}}$  & $\mathbf{d_{JS}}$                      & $\mathbf{d_{C}}$                        \\ 
\midrule
\multirow{10}{*}{\rotatebox[origin=c]{90}{\textbf{MNIST}}}           & \multirow{2}{*}{OH}          & 50                   & $\mathit{0.011 \pm0.006}$         & $\mathit{0.459 \pm0.28}$          & $\mathit{0.266 \pm0.221}$                        & $\mathit{0.811 \pm0.512}$                         \\    
& & 100                   & $\mathit{0.014\pm0.019}$        & $\mathit{0.38\pm0.507}$         & $1.001 \pm0.725$                       & $1.263\pm0.665$                         \\ 
\cline{2-7}
                            & \multirow{2}{*}{SVHN}            & 50                    & $\mathit{0.09\pm0.017}$         & $1.569\pm0.504$        & $6.789 \pm0.924$                       & $12.021\pm1.757$                        \\
                            &                 & 100                   & $0.25\pm0.053$         & $4.702 \pm 1.04$       & $52.349 \pm 3.292$                     & $42.026\pm4.31$                         \\ 
\cline{2-7}
                            & \multirow{2}{*}{TI}              & 50                    & $0.008\pm0.023$       & $1.519\pm0.223$        & $3.663 \pm0.772$                       & $5.512 \pm0.767$                        \\
                            &                 & 100                   & $0.217\pm 0.04$        & $4.3 \pm 0.636$        & $10.305 \pm1.667$                      & $15.18\pm2.698$                         \\ 
\cline{2-7}
                            & \multirow{2}{*}{G}               & 50                    & $0.11 \pm 0.0219$      & $1.958\pm0.534$        & $14.785 \pm1.052$                      & $23.593\pm1.859$                        \\
                            &                 & 100                   & $0.357\pm0.081$        & $5.987 \pm 1.091$      & $52.349\pm4.253$                       & ~$86.21\pm3.471$~                       \\ 
\cline{2-7}
                            & \multirow{2}{*}{SAP}             & 50                    & $0.089\pm0.0311$       & $2.479\pm0.7433$       & $15.116 \pm1.475$                      & $20.151 \pm1.619$                        \\
                            &                 & 100                   & $0.323 \pm 0.07$       & $6.308 \pm 1.366$      & ~$53.397 \pm4.253$                     & $77.456\pm4.474$                       \\ 
\midrule
\multirow{10}{*}{\rotatebox[origin=c]{90}{\textbf{CIFAR-10}}}           & \multirow{2}{*}{OH}       & 50                   & $\mathit{0.056 \pm0.023}$        & $\mathit{0.915 \pm0.934} $         & $\mathit{0.338 \pm0.325}$                       & $0.892 \pm0.402$                         \\         
& & 100                   &   $\mathit{0.061 \pm 0.04}$                     &    $0.769 \pm0.461$                    &$\mathit{0.451 \pm0.41 }$                                        &   $0.648 \pm0.407$                                      \\ 
\cline{2-7}
                            & \multirow{2}{*}{TI}              & 50                    & $0.082 \pm0.037$       & $\mathit{0.928 \pm0.815}$       & $\mathit{0.388 \pm0.243}$                       & $\mathit{0.423 \pm0.362}$                        \\
                            &                 & 100                   & $\mathit{0.056 \pm0.048}$      & $\mathit{0.992 \pm 0.517}$      & $\mathit{0.469 \pm0.426}$                       & $0.415 \pm0.232$                        \\ 
\cline{2-7}
                            & \multirow{2}{*}{SVHN}            & 50                    & $\mathit{0.055 \pm0.032}$       & $\mathit{0.948 \pm0.699}$       & $\mathit{0.665 \pm0.565}$                       & $\mathit{0.414\pm 0.357}$                        \\
                            &                 & 100                   & $\mathit{0.075 \pm 0.036}$      & $1.291 \pm0.925$       & $0.736 \pm 0.658$                      & $0.581 \pm0.343$                        \\ 
\cline{2-7}
                            & \multirow{2}{*}{G}               & 50                    & $\mathit{0.107 \pm0.083}$       & $1.344 \pm 1.156$      & $1.708 \pm0.421$                       & $3.001 \pm0.696$                        \\
                            &                 & 100                   & $0.127 \pm0.087$       & $1.531 \pm0.767$       & $5.855 \pm0.552$                       & $8.703 \pm 0.926$                       \\ 
\cline{2-7}
                            & \multirow{2}{*}{SAP}             & 50                    & $0.1146 \pm0.044$      & $1.854 \pm 0.894$      & $2.299\pm0.691$                        & $2.56 \pm0.762$                         \\
                            &                 & 100                   & $0.208 \pm0.05$      & $5.502 \pm 1.156$      & $8.225 \pm0.866$                       & $9.554 \pm0.489$                        \\ 
\midrule
\multirow{10}{*}{\rotatebox[origin=c]{90}{\textbf{FashionMNIST} }} & \multirow{2}{*}{OH}            & 50                   & $\mathit{0.02 \pm0.012}$        & $\mathit{0.34 \pm0.162 }$         & $\mathit{0.669 \pm0.566}$                       & $\mathit{0.575 \pm0.423}$                         \\    

& & 100                   &   $ 0.059 \pm0.032$ &  $0.801 \pm0.402$ & $0.305 \pm0.237$                   & $0.774 \pm0.343$                  \\ 
\cline{2-7}
                            & \multirow{2}{*}{FP}              & 50                    & $0.105 \pm 0.0526$     & $2.168\pm 0.774$       & $7.263 \pm 0.622$ & $5.305 \pm0.405$   \\
                            &                 & 100                   & $0.195\pm 0.0457$      & $4.819 \pm1.077$       & $9.056 \pm 0.462$ & $11.286 \pm0.751$  \\ 
\cline{2-7}
                            & \multirow{2}{*}{TI}              & 50                    & $\mathit{0.04 \pm0.03}$       & $\mathit{0.798 \pm 0.542}$      & $\mathit{0.897 \pm 0.516}$ & $0.897 \pm0.516$   \\
                            &                 & 100                   & $0.065 \pm0.03$      & $1.66 \pm 0.45$        & $1.4 \pm 0.488$   & $1.912 \pm0.683$    \\ 
\cline{2-7}
                            & \multirow{2}{*}{G}               & 50                    & $\mathit{0.047 \pm 0.03}$       & $\mathit{0.533\pm 0.347}$       & $2.819 \pm 0.703$ & $3.843 \pm0.704$  \\
                            &                 & 100                   & $0.074 \pm0.041$       & $1.325 \pm0.631$       & $9.042 \pm 0.699$  & $15.511 \pm0.445$   \\ 
\cline{2-7}
                            & \multirow{2}{*}{SAP}             & 50                    & $0.036 \pm0.022$       & $0.52 \pm0.303$        & $2.799 \pm 0.497$ & $2.799 \pm0.497$  \\
                            &                 & 100                   & $0.076 \pm0.044$       & $1.411 \pm0.548$       & $8.464 \pm 0.553$  & $8.464 \pm0.553$ \\
\bottomrule
\end{tabular}

\end{tiny}
\end{center}
\end{table}

%% file: tables/correlations.tex
\begin{wraptable}{l}{0.5\textwidth}
\caption{Correlation results for the dissimilarity measures between $S_l$ and $S_u$ with  \gls{OOD} contamination and  \gls{SSDL} accuracy.}
\label{tab:correlations}
\begin{center}
\begin{tiny}

\begin{tabular}{cccccc}
\toprule
\textbf{$\mathbf{S_l}$}       & \textbf{$\mathbf{n_l}$} & \textbf{$\mathbf{d_{\ell_1}}$} & \textbf{$\mathbf{d_{\ell_2}}$} & \textbf{$\mathbf{d_{JS}}$} & \textbf{$\mathbf{d_{C}}$} \\ \hline
\multirow{3}{*}{MNIST}        & 60                      & -0.876                         & -0.898                         & -0.969                     & -0.944                    \\
                              & 100                     & -0.805                         & -0.83                          & -0.786                     & -0.948                    \\
                              & 150                     & -0.794                         & -0.822                         & -0.81                      & -0.944                    \\ \hline
\multirow{3}{*}{CIFAR-10}     & 60                      & -0.823                         & -0.853                         & -0.944                     & -0.921                    \\
                              & 100                     & -0.826                         & -0.878                         & -0.966                     & -0.947                    \\
                              & 150                     & -0.808                         & -0.838                         & -0.952                     & -0.927                    \\ \hline
\multirow{3}{*}{FashionMNIST} & 60                      & -0.2                           & -0.268                         & -0.735                     & -0.789                    \\
                              & 100                     & -0.264                         & -0.326                         & -0.781                     & -0.824                    \\
                              & 150                     & -0.286                         & -0.347                         & -0.785                     & -0.827                    \\  
\bottomrule
\end{tabular}
\vskip -2cm
\end{tiny}
\end{center}
\end{wraptable}